\pdfoutput=1

\documentclass[11pt]{article}

\usepackage[final]{acl}
\usepackage{times}
\usepackage{latexsym}

\usepackage[T1]{fontenc}

\usepackage[utf8]{inputenc}

\usepackage{microtype}

\usepackage{inconsolata}

\usepackage{graphicx}

\title{MoRE: A Mixture of Low-Rank Experts for Adaptive Multi-Task Learning}

\author{
Dacao Zhang\textsuperscript{1}, 
Kun Zhang\textsuperscript{1}\thanks{Corresponding author: Kun Zhang}, 
Shimao Chu\textsuperscript{1}, 
Le Wu\textsuperscript{1}, 
Xin Li\textsuperscript{2,3}, 
Si Wei\textsuperscript{3} 
\\
\textsuperscript{1}School of Computer Science and
Information Engineering, Hefei
University of Technology \\
\textsuperscript{2}School of Information Science and Technology, University of Science and Technology of China \\
\textsuperscript{3}Artificial Intelligence Research Institute, iFLYTEK Company Ltd.\\
  \texttt{\{zhdacao,zhang1028kun,csmao328,lewu.ustc\}@gmail.com, leexin@ustc.edu.cn, siwei@iflytek.com}
  }

\usepackage{algorithm}
\usepackage{algorithmic}
\usepackage{xcolor}
\usepackage{amsmath} 
\usepackage{amssymb}
\usepackage[switch]{lineno}
\usepackage{booktabs}
\usepackage{multirow}
\usepackage{url}
\usepackage{bm}
\usepackage{enumitem}
\usepackage{subcaption}

\usepackage{newfloat}
\usepackage{listings}
\usepackage{tabularx} 
\newcommand{\name}{MoRE}
\newcommand{\la}{LoRA}

\begin{document}
\maketitle

\begin{abstract}
	
	With the rapid development of Large Language Models (LLMs), Parameter-Efficient Fine-Tuning (PEFT) methods have gained significant attention, which aims to achieve efficient fine-tuning of LLMs with fewer parameters. 
	As a representative PEFT method, Low-Rank Adaptation (LoRA) and its variants introduce low-rank matrices to approximate the incremental tuning parameters and achieve impressive performance over multiple scenarios. 
	However, these methods either focus on single-task scenarios or separately train multiple LoRA modules for multi-task scenarios, limiting the efficiency and effectiveness of LoRA in multi-task scenarios.
	To better adapt to multi-task fine-tuning, in this paper, we propose a novel \textbf{M}ixture \textbf{o}f Low-\textbf{R}ank \textbf{E}xperts (\name) for multi-task PEFT. 
	Specifically, instead of using an individual \la~for each task, we align different ranks of \la~module with different tasks, which we named \textit{low-rank experts}. 
	Moreover, we design a novel adaptive rank selector to select the appropriate expert for each task. 
	By jointly training low-rank experts, \name~can enhance the adaptability and efficiency of \la~in multi-task scenarios. 
	Finally, we conduct extensive experiments over multiple multi-task benchmarks along with different LLMs to verify model performance. 
	Experimental results demonstrate that compared to traditional \la~and its variants, \name~significantly improves the performance of LLMs in multi-task scenarios and incurs no additional inference cost. We also release the model and code to facilitate the community\footnote{\url{https://github.com/NLPfreshman0/MoRE}}. 
	
\end{abstract}

\section{Introduction}
\label{s:intro}
\begin{table}[t]
	\centering
	\small
	\begin{tabular}{lcccccc}
		\toprule
		\textbf{Task/Rank} & \textbf{r=1} & \textbf{r=2} & \textbf{r=4} & \textbf{r=8} & \textbf{r=16} & \textbf{r=32} \\ \midrule
		MRPC & \textbf{89.7} & 89.2 & 88.7 & 89.2 & 89.2 & \underline{89.5} \\
		RTE  & 77.6 & 78.7 & \textbf{80.5} & 77.6 & \underline{80.1} & 79.1\\
		SST-2 & 94.4 & \underline{94.6} & \textbf{94.8} & 94.5 & 94.4 & 94.5 \\
		CoLA & 60.9 & 60.0 & 61.9 & \textbf{63.3} & \underline{62.3} & 60.5 \\ \bottomrule
	\end{tabular}
	\caption{LoRA-based Fine-tuning Performance of T5-base with varying ranks on different tasks}
	\label{tab:diff_rank}
\end{table}

Recent advancements in Large Language Models (LLMs) have revolutionized various domains, offering unprecedented performance across numerous tasks~\citep{raffel2020exploring, brown2020language, touvron2023llama}. 
Plenty of tuning strategies are designed to extend the application of LLMs, such as Instruction Tuning~\citep{weifinetuned, zhang2023instruction}, Continual Pre-Training~\citep{kecontinual}, and Parameter-Efficient Fine-Tuning (PEFT)~\citep{houlsby2019parameter, liu2023gpt, lester2021power, hulora}. 
Among these strategies, PEFT has drawn the most attention due to its fewer parameter tuning and lower computational cost. 
As the representative PEFT method, Low-Rank Adaptation (LoRA)~\citep{hulora} introduces low-rank matrices to approximate the incremental tuning parameters and demonstrate good performance in many scenarios, which has become a standard paradigm for LLM fine-tuning and inspired many improvements~\citep{liudora, valipour2023dylora, ding2023sparse}. 

Despite the achieved progress, \la~relies on a fixed and unalterable intrinsic rank, making it not flexible enough in multi-task scenarios. 
Taking Table~\ref{tab:diff_rank} as an example, when dealing with different tasks, \la~requires different ranks to achieve the best performance (e.g., best ranks for MRPC and CoLA tasks are $1$ and $8$). 
Considering the high computational cost and storage cost of LLM fine-tuning, training multiple \la s is sub-optimal for applying LLMs to multi-task scenarios. 
Meanwhile, searching the best rank of \la~during LLM fine-tuning is also time-consuming and computationally expensive~\citep{valipour2023dylora}, which highlights the limitations of a one-size-fits-all approach in LoRA. 
This phenomenon also emphasizes the need for adaptive mechanisms that dynamically adjust ranks based on task requirements.

To overcome the limitations of fixed ranks in \la, one promising direction is to explore adaptive mechanisms. 
For example, DyLoRA~\citep{valipour2023dylora} dynamically trained all ranks during training to avoid separate rank tuning for each task. 
AdaLoRA~\citep{zhang2023adaptive} allocated the parameter budget based on the importance scores of the weight matrices and pruned insignificant singular values to exclude unimportant rank spaces.
SoRA~\citep{ding2023sparse} introduced a trainable gating unit and used proximal gradient descent to optimize the sparsity of the update matrices, thereby dynamically adjusting the intrinsic rank size during training.
While these improvements enable dynamic adjustment of rank space, they are primarily designed for single-task scenarios. They do not consider the distinctions and connections among different tasks in multi-task scenarios, prohibiting the effectiveness of \la~in multi-task scenarios.

In the meantime, there also exist other strategies that try to exploit the connections among different tasks. However, they are still far from satisfied. 
For example, HyperFormer~\citep{mahabadi2021parameter} enhanced adapter-based methods by utilizing a shared hypernetwork to facilitate cross-task knowledge sharing, while incorporating task-specific adapters to tailor the model for individual tasks. 
However, they face limitations due to their inherent performance constraints and additional inference latency. 
Prompt Tuning methods~\citep{vu2022spot, asai2022attempt, wangmultitask} are proposed to use learned prompts on source tasks to initialize the learning of target tasks. 
Despite the effectiveness, these approaches typically require a two-stage training process (i.e., first on the source task and then on the target task), which requires higher data quality and results in training efficiency decrease. 
Meanwhile, parallel LoRA strategies~\citep{wang2023multilora, li2024mixlora, liu2023moelora, huanglorahub} can effectively address the above shortcoming, offering a better adaptability in multi-task scenarios. 
Nonetheless, the usage of parallel LoRA modules increases the overall parameter count and resource consumption, contradicting the original purpose of \la~to reduce the training parameters. 
Thus, one important question should be considered: ``\textbf{How to achieve efficient LLM fine-tuning in multi-task scenarios remains challenging?}''

To this end, in this paper, we design a novel Mixture of Low-Rank Experts (MoRE) for efficient LLM fine-tuning in multi-task scenarios. 
Since different tasks require different ranks of \la, we propose to build connections between the ranks and the tasks in a Mixture-of-Expert (MoE) manner. 
Specifically, we propose to treat each rank in the \la~module as an expert and design a novel \textit{Adaptive Rank Selector}. 
Thus, \textit{the different experts corresponding to different tasks can share common information and maintain distinctive information simultaneously} (i.e., the ranks $r_i$ and $r_j$ can share some common parameters). 
Meanwhile, our proposed selector uses a gating mechanism to select the appropriate rank expert for each task. 
Moreover, to fully exploit the distinctions and connections among different tasks for accurate rank selection, we develop a novel \textit{CL-based Task Embedding} module, which assigns a task embedding to each task and uses a Contrastive Learning~(CL) optimization to ensure the quality of learned task embeddings. 
Furthermore, we incorporate the \textit{Balanced Dataset Sampling strategy} to address the severe dataset imbalance in multi-task scenarios. 
Along this line, \name~can fully exploit the potential of \la~and realize efficient LLM fine-tuning in multi-task scenarios. 
Finally, extensive experiments on multi-task benchmarks demonstrate the efficiency and effectiveness of \name. 

\section{Related Work}
\subsection{Parameter-Efficient Fine-Tuning (PEFT)}
PEFT methods are designed to adapt LLMs to new tasks with minimal additional parameters. 
Representative works include BitFit~\citep{zaken2021bitfit}, Adapters~\citep{houlsby2019parameter}, Prompt Tuning~\citep{liu2023gpt}, Prefix Tuning~\cite{li2021prefix} and Low-Rank Adaptation~(\la)~\citep{hulora}. 
Among these methods, LoRA is the most representative one. 
It introduces trainable low-rank matrices to approximate weight updates, realizing highly efficient fine-tuning with low cost, which has led to various extensions~\citep{kopiczkovera, liudora, valipour2023dylora, zhang2023adaptive, ding2023sparse}. 
For example, VeRA~\citep{kopiczkovera} further reduced the number of trainable parameters in \la~by employing shared low-rank matrices and trainable scaling vectors. DoRA~\citep{liudora} enhanced fine-tuning performance and stability by decomposing the pre-trained weights into magnitude and direction components. 
For flexibility in \la's rank, DyLoRA~\citep{valipour2023dylora} dynamically trained all ranks during training to avoid separate rank tuning for each task. 
AdaLoRA~\citep{zhang2023adaptive} allocated the parameter budget based on the importance scores of the weight matrices and pruned insignificant singular values to exclude unimportant rank spaces. SoRA~\citep{ding2023sparse} introduced a trainable gating unit and used proximal gradient descent to optimize the sparsity of the update matrices, dynamically adjusting the intrinsic rank size during training.

However, \la's fixed-rank constraint limits its flexibility. Although recent works~\citep{valipour2023dylora, zhang2023adaptive} have enhanced LoRA's adaptability, they predominantly address single-task training scenarios. These approaches do not consider multi-task scenarios, where selecting the most suitable rank for different tasks remains an open challenge. This gap underscores the need for more flexible and adaptive methods capable of efficiently handling diverse and concurrent tasks in multi-task learning scenarios.

\subsection{Multi-task learning}
Multi-task learning (MTL) focuses on simultaneously solving multiple related tasks with a single model, which has been studied extensively and offers several advantages~\citep{zhang2021survey,vandenhende2022multi}. 
When integrating with LLMs, new challenges are proposed in MTL scenarios, such as task conflicts, balancing task weights, and training resource demands~\citep{chen2021multi, kollias2024distribution}. 
Many methods are developed to tackle these problems. 
E.g., HyperFormer~\citep{mahabadi2021parameter} enhanced Adapter-based methods with a shared hypernetwork for cross-task knowledge sharing; SPoT~\citep{vu2022spot} adapted learned prompts for target tasks to improve performance; ATTEMPT~\citep{asai2022attempt} merged source and target prompts using an attention mechanism; and MPT~\citep{wangmultitask} used prompt decomposition and knowledge distillation for creating transferable prompts with low-rank modifications for task specificity.

Moreover, LoRA-based enhancements like MultiLoRA~\citep{wang2023multilora}, MixLoRA~\citep{li2024mixlora}, and MOELoRA~\citep{liu2023moelora} employ multiple parallel \la~modules or experts with gating mechanisms to manage task-shared and specific knowledge. However, these methods often increase trainable parameters, impacting training efficiency, and do not always accommodate the different rank needs of tasks.
\section{Preliminary}
\label{s:preliminary}

\subsection{Problem Definition}
In multi-task learning scenarios, the objective is to concurrently learn multiple tasks, each characterized by potentially diverse data distributions and goals. 
Formally, we consider a set of tasks $T = \{\mathcal{T}_1, \mathcal{T}_2, \ldots, \mathcal{T}_T\}$, where each task $\mathcal{T}_t$ is associated with a dataset $\mathcal{D}_t = \{(x_i^t, y_i^t)\}_{i=1}^{N_t}$ comprising $N_t$ input-output pairs. $x_i^t$ denotes the input data and $y_i^t$ denotes the label or output for task $\mathcal{T}_t$. 
The target is to learn a shared model $F$ to satisfy the requirements of different simultaneously.

\begin{figure*}
	\begin{center}
		\includegraphics[width=0.86\textwidth]{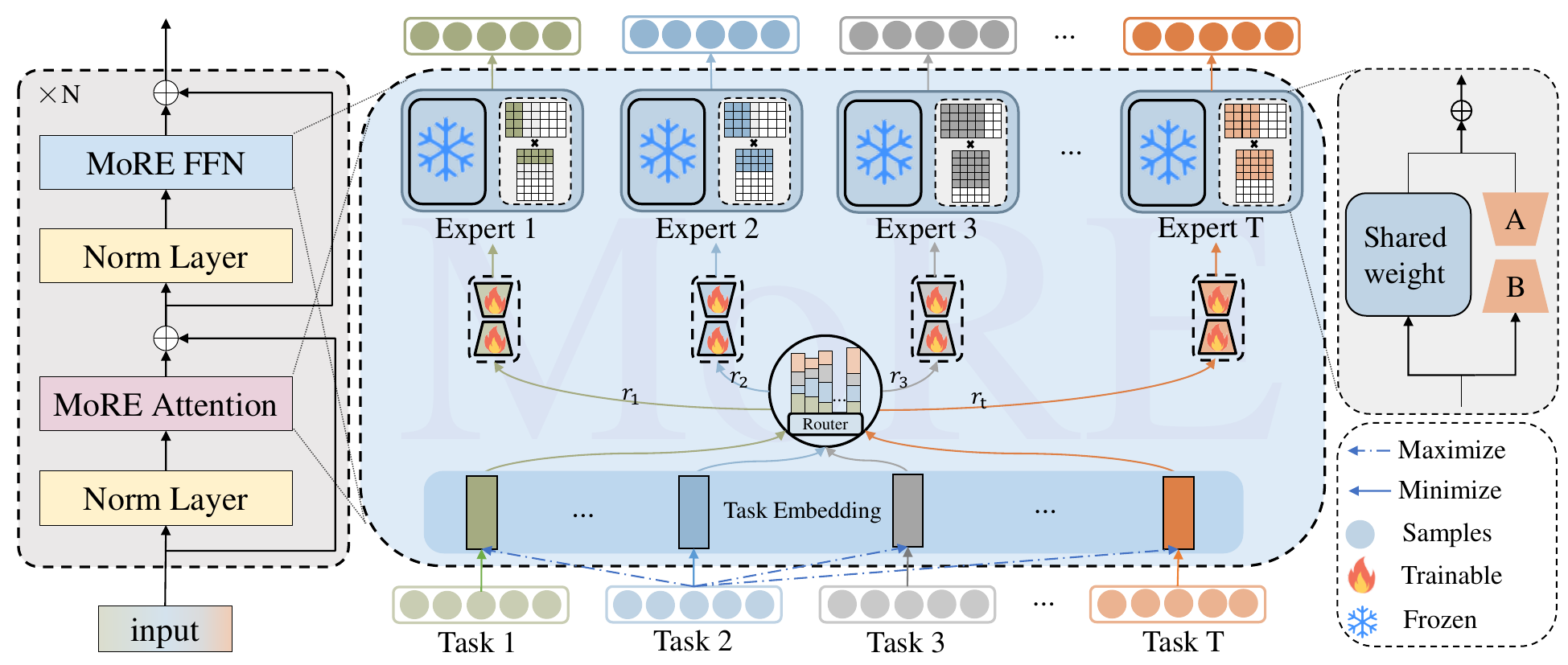}
	\end{center}
	\caption{The overall framework of our proposed MoRE.}
	\label{f:framework}
\end{figure*}


\subsection{LoRA: Low-Rank Adaptation}
LoRA \citep{hulora} is designed to reduce the computational cost and memory footprint of adapting LLMs by introducing low-rank updates to the weight matrices. Given the original weight matrix $\bm{W}_0 \in \mathbb{R}^{m \times d}$, LoRA approximates the weight update $\Delta \bm{W} = \bm{B}\bm{A}$, where $\bm{A} \in \mathbb{R}^{r \times d}$ and $\bm{B} \in \mathbb{R}^{m \times r}$ are low-rank matrices, and $r \ll \min(m, d)$ is the rank. The modified forward pass becomes:
{
	\begin{equation}
		\bm{h} = \bm{W}_0\bm{x} + \bm{B}\bm{A}\bm{x}.
	\end{equation}
}




However, this process highly depends on the pre-defined rank $r$, which is time-consuming and computationally expensive to search. And this problem will be amplified in multi-task scenarios, limiting the potential of \la.
Thus, \textit{How to use LoRA to achieve efficient LLM fine-tuning in multi-task scenarios} is the main focus of our paper. 


\section{Mixture of Low-Rank Experts}
\label{s:model}
To tackle the inefficient problem of LoRA in multi-task scenarios, we propose a novel Mixture of Low-Rank Experts (\name). 
The cores lie in \textit{how to learn experts} and \textit{how to select them}. 
As illustrated in Figure~\ref{f:framework}, we focus on parameters in attention layer and FFN layer of the Transformer block. 
We first assign a task embedding for each task to describe the abstract task characteristics. 
Then, based on the task embedding, we design a novel \textit{adaptive rank selector} to select the appropriate rank for each task, term as the rank expert. 
Finally, we incorporate contrastive learning to ensure the quality of learned task embedding and design a \textit{Balanced Data Sampling strategy} to stabilize the learning process for better multi-task learning. 
Next, we will introduce each part in detail.



\subsection{Task Embedding}
\label{s:task-embedding}
Existing multi-task learning methods focus on mining useful information from task data and transferring knowledge from one task to another. 
Despite the progress, they are still weak at sharing common information among tasks and distinguishing specific information aligning with each task. 
This shortcoming will prohibit the efficiency of PEFT methods when using them to tune LLMs in multi-task scenarios. 
Therefore, we propose using task embeddings to represent different tasks so that task characteristics can be summarized comprehensively. This operation is also the precondition of our designed rank expert for measuring the connections and distinctions among different tasks.

Specifically, we use matrix $\bm{E}=\{\bm{e}_1, \bm{e}_2, ..., \bm{e}_l\}$ to denote all tasks, where $\bm{e}_i$ represents the $i^{th}$ task in the multi-task scenarios. 
Then, we leverage Kaiming Initialization to initialize them and learn precise $\bm{E}$ during model training. 
Since there is no supervised signal for $\bm{E}$, we design a Contrastive Learning (CL) based optimization target to learn them, which will be introduced in Section~\ref{s:sampling-learning}.

\subsection{Adaptive Rank Selector}
\label{s:rank-selector}
As illustrated in Section~\ref{s:intro}, LoRA and its typical variances usually have a pre-defined fixed rank $r$. 
However, different tasks may benefit from different ranks depending on their complexity and data distributions~\citep{valipour2023dylora, ding2023sparse}. 
Searching the best rank is time-consuming and computationally expensive. 
Meanwhile, training parallel \la~modules or multiple \la s when applying LLMs to multi-task scenarios will amplify the problem and prohibit the effectiveness, causing high computational and storage costs. 
Therefore, we employ Mixture-of-Experts~(MoE) framework and design a novel \textit{Adaptive Rank Selector}. 

Different from previous work that treated the entire LoRA module as an expert, we propose to treat the rank $r$ as the expert and use one \la~to realize LLM fine-tuning in multi-task scenarios. 
Assuming the selected rank of \la~is $r$, the rank expert can be selected within the range $[1, r]$. 
\textit{Along this line, different experts can share common information at the overlap part in the learned metrics} (i.e., $\bm{A}$ and $\bm{B}$) \textit{and align specific information corresponding to each task at the non-overlap part.} 
Formally, we use the learned task embedding $\bm{e}_t$ to select the appropriate rank from the \la~module and 
leverage a gating network $\textit{G}(\cdot)$ to guarantee the quality of the selection. 
Let $\{1, 2, \ldots, r\}$ be the set of experts' ranks. 
For task $\mathcal{T}_t$, $\textit{G}(\cdot)$ takes $\mathbf{e}_t$ as input and outputs a probability distribution over rank experts as follows:
\begin{equation}
	\label{eq:gating}
	\mathbf{p}_t = \textit{G}(\mathbf{e}_t) = \text{softmax}(\mathbf{W}_g \mathbf{e}_t + \mathbf{b}_g),
\end{equation}
where $\{\mathbf{W}_g, \mathbf{b}_g\}$ are learnable parameters. 
The probability distribution $\mathbf{p}_t \in \mathbb{R}^r$ indicates the relevance of each rank to task $\mathcal{T}_t$. During the forward pass, we select the rank with the highest probability and use the selected rank to truncate LoRA module for rank expert construction. Then, \name~uses LoRA paradigm to realize the fine-tuning:
{
	\begin{equation}
		\label{eq:expert-construct}
		\begin{gathered}
			r_t = \arg\max \mathbf{p}_t, \\
			\bm{h} = \mathbf{W_0}x + \mathbf{B_tA_t}x, \\ 
			\mathbf{A_t} = \mathbf{A}[:r_t,:], \quad \mathbf{B_t} = \mathbf{B}[:,:r_t]. \\
		\end{gathered}
	\end{equation}
}


One step further, during backward pass, the $\arg\max$ in Eq.(\ref{eq:gating}) is non-differentiable, causing $\textit{G}(\cdot)$ unable to be learned. 
Thus, we incorporate Straight-Through Estimator (STE)~\citep{bengio2013estimating} technique to address this issue. 
We use STE to calculate the approximate gradient to allow the gradient to propagate back to $\textit{G}(\cdot)$ correctly:
\begin{equation}
	\text{Ste}(\mathbf{p}_t) = \mathbf{p}_t + sg[one\_hot(\mathbf{p}_t) - \mathbf{p}_t], \\
	\label{ste}
\end{equation}
where $one\_hot(\cdot)$ is used to convert a vector into its one-hot version. $sg(\cdot)$ stands for stop gradient. Then, we modify the forward process in Eq.(\ref{eq:expert-construct}) as:
{
	\begin{equation}
		\mathbf{h} =  \mathbf{W_0}x + \text{Ste}(\mathbf{p}_t)[r_t] \cdot \mathbf{B_tA_t}x. \\
	\end{equation}
}

Thus, Adaptive Rank Selector module can realize a precise selection of rank experts.
Furthermore, since \name~uses the overlap part among LoRA metrics to share the common information across different tasks, the lower part will be updated more frequently during fine-tuning. Thus, its learning rate should be small for a slow and stable updating. 
To realize this goal, we perform a linear scaling on its weights for the balance:
\begin{equation}
	\label{eq:linear}
	\mathbf{h} =  \mathbf{W_0}x + \text{Ste}(\mathbf{p}_t)[r_t] \cdot \frac{r_t}{|T|} \mathbf{B_tA_t}x, \\
\end{equation}
where $|T|$ is the total number of tasks. To verify the effectiveness of this design, we also conducted an ablation study on this operation in Section \ref{s:ablation}. 

We have to note that \name~is largely different from training multiple \la~with $r=1$. The latter still use the parallel paradigm and does not consider the connections and distinctions among different tasks. 
In contrast, \name~uses adaptive rank selector to dynamically assign suitable rank for different tasks (i.e., The more similar the tasks are, the closer the expert rank is and vice versa). 


\subsection{Balanced Data Sampling and CL-based Optimization}
\label{s:sampling-learning}
\noindent \textbf{Balanced Data Sampling. }
In multi-task scenarios, data distributions of different tasks are also essential for LLM fine-tuning. 
For instance, in GLUE benchmark~\citep{wang2018glue}, MNLI and RTE datasets have proportionally disparate data distributions (i.e., $392,000$ v.s. $2,500$ examples).
If this attribute is not considered when fine-tuning LLMs in multi-task scenarios, it is obvious that fine-tuned LLMs will underfit the task with smaller datasets. 

In response, we propose a simple but effective Balanced Dataset Sampling strategy to ensure each dataset contributes proportionally during the fine-tuning process, regardless of its size. Specifically, we assign a sampling weight $\phi_t$ to each dataset $\mathcal{D}_t$, which is inversely proportional to its size:

{\small
	\begin{equation}
		\label{eq:bds}
		\begin{gathered}
			{\Phi} = [\phi_1, \phi_2, ..., \phi_T], \quad  
			\phi_t = \exp\left(\frac{|\mathcal{D}_t|}{\sum_{i=1}^T |\mathcal{D}_i|}\right), \\
			D_t = \textit{Sampling}(D, \Phi),
		\end{gathered}
	\end{equation}
}
where $\textit{Sampling}(D, \Phi)$ denotes sampling a subset from all datasets $D$ with the distribution $\Phi$. 
$|\mathcal{D}_t|$ is the size of dataset $\mathcal{D}_t$. 
This dynamic sampling strategy helps to balance the contributions of different datasets, thereby reducing the risk of underfitting smaller datasets and improving the overall performance of the multi-task training. 
\setlength{\tabcolsep}{4pt}
\begin{table*}
	\scriptsize 
	\centering
	\begin{tabular*}{\textwidth}{@{\extracolsep{\fill}}lcccccccccc}
		\toprule
		\textbf{Methods} & \textbf{params/task} & MNLI & QQP & QNLI & SST-2 & STS-B & MRPC & RTE & CoLA & AVG \\
		\midrule
		Finetuning & 28M & 85.7 & \textbf{91.1} & 92.0 & 92.5 & 88.8 & \underline{90.2} & 75.4 & 54.9 & 83.8 \\
		Adapters & 1.8M & \textbf{86.3} & 90.5 & \underline{93.2} & 93.0 & 89.9 & \underline{90.2} & 70.3 & 61.5 & 84.4 \\
		PT & 9.6k & 85.6 & \underline{90.6} & \underline{93.2} & 93.9 & 89.9 & 86.3 & 67.6 & 55.3 & 82.8 \\
		$LoRA_{r=8}$ & 0.39M & 85.8 & 89.2 & 93.1 & 93.2 & \underline{90.4} & 89.9 & 76.3 & 62.8 & 85.1 \\
		$LoRA_{r=16}$ & 0.78M & 84.9 & 89.6 & 93.0 & 93.7 & \underline{90.4} & 88.7 & 80.6 & 63.9 & 85.6 \\
		\midrule
		HyperFomer & 638K & 85.7 & 90.0 & 93.0 & 94.0 & 89.7 & 87.2 & 75.4 & 63.7 & 84.8 \\
		MPT & 10.5K & 84.3 & 90.0 & 93.0 & 93.3 & \underline{90.4} & 89.2 & \underline{82.7} & 63.5 & 85.8 \\
		MultiLoRA & 1.56M & 85.9 & 89.7 & 92.8 & \textbf{94.5} & 89.8 & 88.2 & 80.6 & 66.9 & 86.0 \\
		MixLoRA & 1.49M & 85.8 & 90.0 & 92.9 & 93.7 & 90.3 & 89.2 & 78.4 & 67.2 & 85.9 \\
		MOELoRA & 0.78M & \textbf{86.3} & 90.1 & \underline{93.2} & \underline{94.2} & 90.0 & 89.7 & 81.3 & \underline{68.4} & \underline{86.7} \\
		\midrule
		\name & 0.78M & \underline{86.2} & 90.0 & \textbf{93.4} & 93.7 & \textbf{90.7} & \textbf{91.2} & \textbf{83.5} & \textbf{69.9} & \textbf{87.3} \\
		\midrule
		LLaMA2-LoRA & 2.5M & 86.9 & \underline{88.6} & 93.5 & 96.2 & 90.2 & \textbf{92.6} & \underline{89.2} & 65.0 & 87.8 \\
		LLaMA2-MultiLoRA & 10M & \underline{87.6} & 85.0 & 93.4 & \underline{96.7} & \underline{92.2} & 88.7 & 87.8 & \underline{72.4} & \underline{88.0} \\
		LLaMA2-MixLoRA & 12.2M & 86.8 & 88.1 & \underline{93.6} & 96.0 & 91.3 & 88.2 & 87.1 & \textbf{73.2} & \underline{88.0} \\
		LLaMA2-MOELoRA & 5M & 87.0 & 87.6 & 91.4 & 96.3 & \textbf{92.4} & \underline{91.2} & 87.8 & 64.4 & 87.3 \\
		\midrule
		LLaMA2-MoRE & 5M & \textbf{89.4} & \textbf{89.0} & \textbf{94.4} & \textbf{96.9} & \underline{92.2} & 89.2 & \textbf{92.1} & 66.9 & \textbf{88.8} \\
		\bottomrule
	\end{tabular*}
	\caption{Performance on GLUE benchmark. For STS-B, we report Pearson correlation coefficients. For CoLA, we report Matthews correlation. For all other tasks, we report Accuracy. \textbf{Bold} and \underline{underlined} fonts indicate the best and the second-best results.}
	\label{tab:overall_results}
\end{table*}

\begin{table}
	\centering
	\scriptsize  
	\setlength{\tabcolsep}{2pt} 
	\begin{tabular}{cccccccc}
		\toprule
		\textbf{Methods} & \textbf{params/task} & \textbf{BoolQ} & \textbf{PIQA} & \textbf{OBQA} & \textbf{ARC-E} & \textbf{ARC-C} & \textbf{AVG} \\
		\midrule
		LoRA & 2.5M & 80.9 & 77.7 & 79.0 & 83.7 & \underline{76.9} & 79.6 \\
		MultiLoRA & 10M & 76.5 & 72.9 & 68.2 & 81.6 & 61.9 & 72.2 \\
		MixLoRA & 12.2M & \underline{84.3} & 79.5 & \underline{82.6} & \textbf{86.8} & 76.3 & 81.9 \\
		MOELoRA & 4.5M & 84.0 & \underline{79.9} & 81.8 & \textbf{86.8} & \textbf{77.3} & \underline{82.0} \\
		\midrule
		\name & 4.5M & \textbf{87.2} & \textbf{82.3} & \textbf{83.0} & \underline{86.7} & 74.2 & \textbf{82.7} \\
		\bottomrule
	\end{tabular}
	\caption{Accuracy of all methods on Commonsense Reasoning tasks. The backbone is Llama2-7B.}
	\label{tab:commonsense}
\end{table}

\noindent \textbf{CL-based Optimization. }
As mentioned in Section~\ref{s:task-embedding}, there is no supervised signal for task embedding learning. 
Thus, one important question should be considered: ``\textit{How to ensure the task characteristics and task distinguishability of the learned task embedding without annotation requirements?}'' 
In response, we propose to leverage CL to ensure the quality of learned task embeddings. 
Consider a batch $\mathcal{B}$ of samples, where all samples in $\mathcal{B}$ belong to the same task $\mathcal{T}_t$. Let $\{ \mathbf{x}_i \}_{i=1}^N$ be the set of $N$ samples in $\mathcal{B}$, and let $\mathbf{h}_i$ be the representation of sample $\mathbf{x}_i$ obtained from the model. The task embedding for task $\mathcal{T}_t$ is denoted as $\mathbf{e}_t$.
The optimization target can be formulated as follows:

{
	\small
	\begin{equation}
		\label{eq:cl-target}
		\mathcal{L}_{\textit{con}} = \frac{1}{N} \sum_{i=1}^N \left[ \log \frac{\exp\left(\frac{\textit{sim}(\mathbf{h}_i, \mathbf{e}_t)}{\tau}\right)}{\sum_{k=1}^T \exp\left(\frac{\textit{sim}(\mathbf{h}_i, \mathbf{e}_k)}{\tau}\right)} \right], \\
	\end{equation}
}
where $\textit{sim}(\cdot, \cdot)$ denotes a similarity measure, such as the dot product or cosine similarity, and $T$ is the total number of tasks. $\tau$ is the temperature. 
$\bm{e}_t$ and $\bm{e}_k$ are the $t^{th}$ and $k^{th}$ tasks ($t\neq k$). 
By using Eq.(\ref{eq:cl-target}), we can measure the connection between task embedding $\bm{e}_t$ and its data samples $\{ \mathbf{x}_i \}_{i=1}^N$. 
Since each data sample is close to the corresponding task embedding, we can conclude the learned task embeddings can be used to describe task characteristics, which is also supported by experimental results in Section~\ref{s:expert-selection}.

Besides using contrastive loss to learn task embeddings, we also select generation loss \( \mathcal{L}_{\textit{gen}} \) to measure the discrepancy between the generated sequences and target sequences. 
Let $\bm{y}$ and $\hat{\bm{y}}$ be target sequence and generation, \( \mathcal{L}_{\textit{gen}} \) can be formulated with the cross-entropy loss: 
{
	\begin{equation}
		\mathcal{L}_{\textit{gen}} = - \sum_{t=1}^{T} y_t \log \hat{y}_t. \\
		\label{gen_loss}
	\end{equation}
}

\noindent Then, we leverage a hyperparameter $\lambda$ to balance the contributions of the generation loss and the contrastive loss, and formulate the overall optimization target of \name~as follows:
{
	\begin{equation}
		\label{eq:total_loss}
		\mathcal{L} = \mathcal{L}_{\textit{gen}} + \lambda \mathcal{L}_{\textit{con}}. \\
	\end{equation}
}


\paragraph{Discussion.}
Compared with existing methods, \name~has the following properties. 
1) We propose to treat different rank $r$ in one \la~as experts, and design an adaptive rank selector to select suitable rank experts for different tasks, which can effectively measure the connections and distinctions among different tasks;
2) We use task embeddings to accurately describe the task characteristics with a CL optimization; 
3) We also consider task data distributions and design a simple but effective Balanced Data Sampling strategy to ensure the capability of fine-tuned LLMs on different tasks.

\section{Expertments}
\label{s:experiments}
\setlength{\tabcolsep}{3pt}
\begin{table*}
	\scriptsize 
	\centering
	\begin{tabular*}{\textwidth}{@{\extracolsep{\fill}} lccccccccc}
		\toprule
		\textbf{Task} & \textbf{k-shot} & \textbf{Finetuning} & \textbf{LoRA}
		& \textbf{HyperFomer} & \textbf{MPT} & \textbf{MultiLoRA} & \textbf{MixLoRA} & \textbf{MOELoRA} & \textbf{MoRE} \\
		\midrule
		\multirow{3}{*}{BoolQ} & 4 & 50.5 & 64.2 & 48.0 & 62.2 & \textbf{65.2} & 62.8 & 64.0 & \underline{64.6} \\
		& 16 & 56.5 & \underline{66.1} & 50.2 & 63.3 & 65.8 & 64.4 & 64.8 & \textbf{66.2} \\
		& 32 & 58.4 & 67.4 & 58.3 & \textbf{68.9} & 67.6 & 66.2 & 65.7 & \underline{67.9} \\
		\midrule
		\multirow{3}{*}{CB}    & 4 & 57.7 & 84.3 & 60.7 & 73.6 & 85.0 & \textbf{86.6} & 85.4 & \underline{85.7} \\
		& 16 & 77.0 & 85.7 & 76.3 & 78.6 & 85.7 & \textbf{86.4} & \underline{86.3} & \textbf{86.4} \\
		& 32 & 80.0 & 87.1 & 81.4 & 82.1 & 86.6 & \textbf{89.3} & 88.3 & \underline{88.6} \\
		\midrule
		\multirow{3}{*}{SciTail} & 4 & 79.6 & 80.8 & \underline{82.0} & 80.2 & 78.1 & 77.5 & 80.4 & \textbf{83.8} \\
		& 16 & 80.0 & 84.0 & 86.5 & \textbf{87.3} & 81.7 & 82.4 & 83.1 & \underline{86.7} \\
		& 32 & 81.9 & 85.3 & 85.8 & \underline{86.3} & 83.6 & 83.3 & 84.5 & \textbf{87.4} \\
		\bottomrule
	\end{tabular*}
	\caption{Few-shot domain transfer results (Accuracy) of T5-base models fine-tuned on GLUE averaged across 5 seeds. \textbf{Bold} and \underline{underlined} fonts indicate the best and the second-best results.}
	\label{tab:fewshot_results}
\end{table*}

\subsection{Experimental Setup}
\label{s:setup}
\textbf{Datasets.}
We evaluated the model using GLUE benchmark~\citep{wang2018glue} to assess various natural language understanding tasks. Additionally, we included datasets like BoolQ~\citep{clark2019boolq}, PIQA~\citep{bisk2020piqa}, OBQA~\citep{mihaylov2018can}, and ARC~\citep{clark2018think} to test commonsense reasoning abilities. Moreover, we select SciTail~\citep{khot2018scitail}, BoolQ~\citep{clark2019boolq}, and CB~\citep{de2019commitmentbank} datasets to evaluate model robustness and generalization in few-shot learning scenarios.
We also report performance on generation tasks in Appendix~\ref{s:nlg}. 

\noindent \textbf{Baselines.}
The following baselines are selected: 
1) {Full fine-tuning (FT)}, 
2) {Vanilla Adapter},
3) {Vanilla prompt tuning (PT)},
4) {Vanilla LoRA}.
We also select the following advanced multi-task PEFT baselines:
1) HyperFomer,
2) MPT,
3) MultiLoRA,
4) MixLoRA,
5) MOELoRA.
All methods are tuned based on reported settings for a fair comparison. 



\noindent \textbf{ Implementation.} 
We utilized LLaMA2-7B and T5-base as backbones with the AdamW optimizer. The learning rate was set to $3\times 10^{-4}$, applying a linear decay with a warm-up phase over the first 500 steps. The training was conducted over 5 epochs with a batch size of 32 and a maximum input sequence length of 128 tokens. Parameter $\lambda$ was set to 0.1 and the softmax temperature $\tau$ to 0.05. For few-shot domain transfer, we initialized with the best checkpoint from GLUE task training, sharing task embeddings for similar tasks. T5-base was trained on two NVIDIA RTX 4090 GPUs, and LLaMA2-7B on four NVIDIA Tesla A100 GPUs.




\begin{table}[t]
	\centering
	\small
	\begin{tabular}{lc}
		\toprule
		\textbf{Conditions} & \textbf{GLUE Avg.} \\
		\midrule
		\name & \textbf{87.3} \\
		w/o Linear Scaling & \underline{87.0} \\
		w/o Task Embeddings & 86.1 \\
		w/o CL optimization & 86.3 \\
		w/o STE & 86.4 \\
		w/ Subset Experts & 86.2 \\
		w/ Random Sample & 86.2 \\
		\bottomrule
	\end{tabular}
	\caption{Ablation study results (Average Results on GLUE benchmark) of \name.}
	\label{tab:ablation_study}
\end{table}

\subsection{Overall Performance}
\textbf{Performance on GLUE Benchmark and Commonsense Reasoning.}
Tables~\ref{tab:overall_results} and~\ref{tab:commonsense} show that \name~excels in multi-task scenarios with few fine-tuned parameters, outperforming LoRA implementations. By using task-specific embeddings, 
\name~efficiently manages task information, enhancing performance without excessive parameter tuning. This efficiency extends to large models like LLaMA2-7B, significantly boosting performance. 

In contrast, PEFT baselines struggle with small datasets due to a lack of shared knowledge integration and higher data demands for training, leading to suboptimal results. Multi-task baselines, while considering shared knowledge, fail to appropriately differentiate task nuances, resulting in inferior performance compared to \name. Approaches like MultiLoRA and MixLoRA improve performance but lack task-aware mechanisms and specific rank allocations, limiting their effectiveness. Additionally, they tend to have more trainable parameters. 

In commonsense reasoning tasks, \name~also leads with the highest accuracy, proving its robustness across different scenarios. This suggests that \name~can effectively handle the nuanced requirements of commonsense reasoning than simpler ensemble approaches like MixLoRA or MoELoRA.

\textbf{Performance on Few-shot Domain Transfer.}
We conducted few-shot domain transfer experiments to test the efficiency of \name, with results detailed in Table~\ref{tab:fewshot_results}. \name~consistently performs well across various datasets and few-shot settings, demonstrating its capability to efficiently share and distinguish task-specific information for effective transfer learning. Traditional fine-tuning methods, including HyperFormer and MPT, require more training data to achieve better results. LoRA-based multi-task methods, in contrast, do not outperform standard LoRA implementations in these settings, likely due to difficulties in rank allocation and parameter adaptation with few samples. This underscores the challenges of few-shot learning and highlights the effectiveness of \name~ in achieving better generalization across different domains.

\begin{figure*}
\begin{center}
	\includegraphics[width=1.0\textwidth]{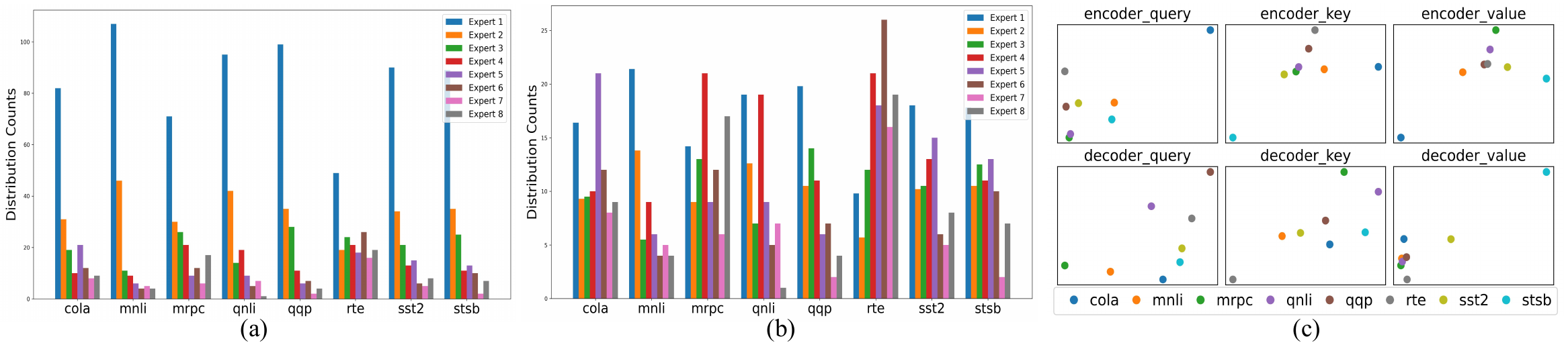}
\end{center}
\caption{(a)-(b) The distribution of expert allocation. (c) Visualization of the task embeddings.}
\label{fig:visual}
\end{figure*}

\begin{figure}[t]
\centering
\includegraphics[width=0.4\textwidth]{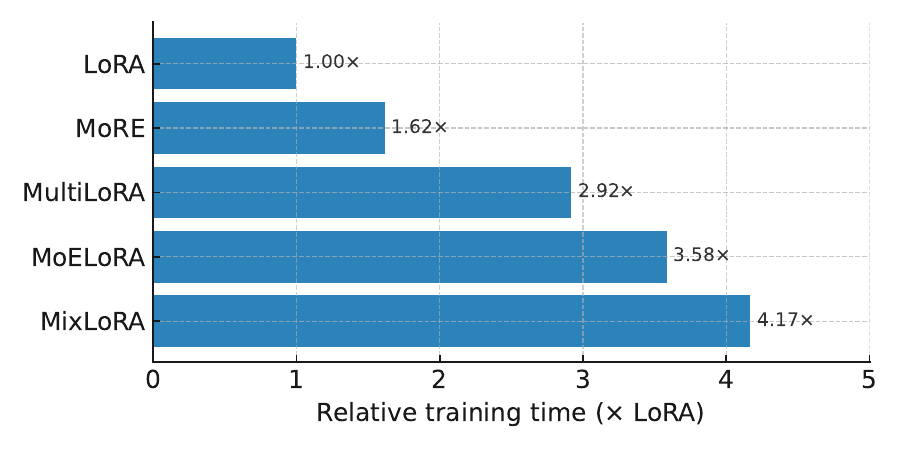}
\caption{Relative Training Speed of Different Parameter-Efficient Fine-Tuning Methods.}
\label{fig:speed}
\end{figure}

\subsection{Detailed Analysis}
\label{s:expert-selection}
\textbf{Low-Rank Expert Allocation. }
We analyzed the expert distribution across all layers for each task after fine-tuning, as illustrated in Figure \ref{fig:visual}(a). Most tasks predominantly utilized experts ranked 1, 2, or 3, suggesting parameter redundancy in higher ranks within LoRA modules during fine-tuning. This aligns with our design where \name~leverages lower-rank experts to share common information across tasks. To further clarify task dependencies on different ranks, we scaled down the influence of experts 1-3, as shown in Figure~\ref{fig:visual}(b). The analysis revealed specific dependencies, such as MRPC on expert 4, confirming \name's ability to effectively assign appropriate experts to tasks, thereby enhancing performance in multi-task environments.

\noindent \textbf{Visualization of Task Embeddings. }
In Section~\ref{s:task-embedding}, we discuss the crucial role of task embeddings in selecting rank experts for \name. We visualized these embeddings using PCA from the self-attention module's final layer, as shown in Figure~\ref{fig:visual}(c). The results reveal clear clustering patterns among similar tasks, like MRPC and QNLI, and significant separations for distinct tasks, particularly STSB and CoLA. This clustering aligns with the nature of the tasks, with STSB focusing on similarity computation, differing fundamentally from classification tasks. These insights confirm the effectiveness of \name~in using task embeddings to differentiate and link tasks, enhancing expert selection and the overall performance of \name. We also provide more examples in Appendix~\ref{s:add_vis}.

\noindent \textbf{Training Speed Analysis. }
Figure~\ref{fig:speed} shows the training time per step, normalised to the LoRA baseline ($1\times$).
Our proposed \name, is only ${\sim}1.6\times$ slower than LoRA.
Mixture-style adapters are much slower.
MultiLoRA and MoELoRA need about $2.9\times$ and $3.6\times$ more time because they must manage several experts.
MixLoRA is the slowest at roughly $4.2\times$; its gated routing increases GPU control-flow divergence.
These results confirm that simpler designs run faster.
\name{} delivers the best overall trade-off: high accuracy with minimal speed cost.

\begin{figure}[t]
\centering
\includegraphics[width=0.4\textwidth]{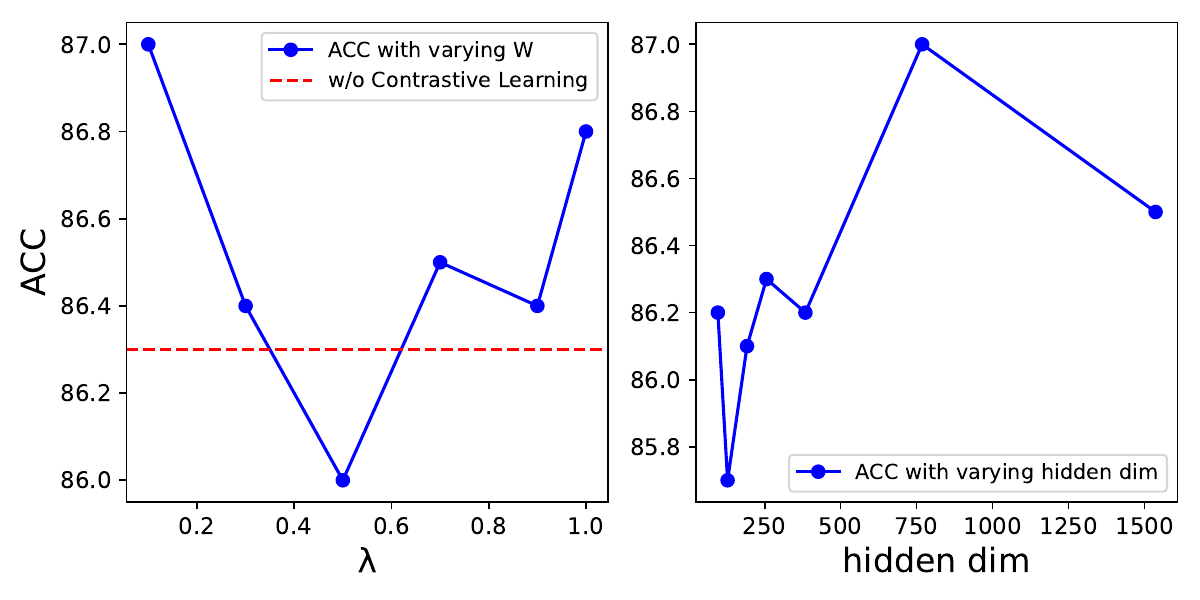}
\caption{Parameter Sensitivity Test on $\lambda$ in Eq.(\ref{eq:total_loss}) and hidden dimension of task embedding.}
\label{fig:params}
\end{figure}

\subsection{Ablation Study and Parameter Analysis}
\label{s:ablation}
\textbf{Ablation Study. } 
We conducted an ablation study for \name, with results detailed in Table~\ref{tab:ablation_study}. The study highlights significant performance declines when task-specific embeddings or contrastive optimization are omitted, confirming their crucial roles. Removing STE and using soft expert selection also drastically reduces performance. Using random sampling reduced the results, supporting the effectiveness of our balanced sampling strategy. Flexible rank selection by allowing any subset as experts led to worse outcomes, likely because foundational ranks typically harbor broader, shareable knowledge crucial for task performance. Minor drops in performance without linear scaling indicate its role in preventing overfitting. 

\noindent \textbf{Parameter Sensitivity Test. }
We analyzed the impact of two key hyperparameters on model performance: the $\lambda$ value and the dimension of task embeddings, with results shown in Figure~\ref{fig:params}. We observed that increasing $\lambda$ initially lowers model performance due to oscillations in contrastive loss across diverse datasets, stabilizing at $\lambda=0.1$ for optimal performance. Regarding task embedding dimensions, performance improves with dimension increases up to a point before declining. Smaller dimensions fail to capture complex task details, while larger dimensions require excessive data for effective training and are cumbersome when calculating sample similarities. Consequently, we selected a task embedding dimension of 768.

\begin{table}[t]
\centering
\scriptsize
\begin{tabular}{lc}
	\toprule
	\textbf{Method} & \textbf{Parameter} \\
	\midrule
	LoRA &  $6Lr(m + d)$ \\
	MultiLoRA & $6nLr(m + d) + 6Ld$ \\
	MixLoRA & $2nLr(m + d) + 2Lnm$ \\
	MOELoRA & $6Lr(m + d)+ 6Lh(n + T)$ \\
	\name & $6Lr(m + d) + 6Lh(r + T)$ \\
	\bottomrule
\end{tabular}
\caption{Parameter sizes of different methods based on model layers ($L$), LoRA rank ($r$), model dimensions ($m$ and $d$), number of parallel LoRA modules ($n$), task numbers ($T$), and task embedding dimension ($h$).}
\label{tab:methods_params}
\end{table}

\noindent \textbf{Parameter Efficiency. }
To analyze the model complexity, we count the number of tuning parameters of different \la-based methods and report results in Table~\ref{tab:methods_params}. 
Compared with \la,  the added parameter number of \name~is $6Lh(r+T)$, including extra task embeddings ($Th$ for a single \la~module) and adaptive rank selector ($rh$ for a single \la~module). 
Compared with multi-task baselines, \name~is more efficient. 
Moreover, once \name~are trained, we can construct a mapping from tasks to experts during inference, thereby reducing the parameter count to be consistent with \la. 
We also provide detailed model complexity analysis in Appendix~\ref{s:parameter-count}.

\section{Conclusion}
In this paper, we addressed the inefficiencies of existing PEFT methods that often require too many tuning parameters for multi-task fine-tuning. We introduced a novel approach, \name, which optimizes the use of low-rank parameters in \la~modules by treating each as a specialized expert. This strategy allows for sharing common information through lower ranks while emphasizing task-specific details through higher ranks. We enhanced the selection of these expert ranks using task embeddings and supported fine-tuning with techniques like CL-based optimization and Balanced Dataset Sampling. Our extensive testing on the GLUE benchmark shows substantial improvements and promising transfer learning capabilities.


\section{Limitations}
\label{s:limination}
Despite the achieved progress, our proposed \name~still has some limitations. 
First, due to GPU device limitations, we do not apply \name~to larger LLMs, such as 13B, 75B, etc;
Second, though we have made an early attempt on generation tasks in Appendix \ref{s:nlg}, detailed experiments are needed to better verify the effectiveness of \name.
Finally, since our approach is based on the MoE structure, which cannot be merged with the original model, it results in latency during inference. Although \name~ has significantly improved efficiency compared to traditional MoE approaches, further improvement is still worth exploring.

\section{Acknowledgements}
This research was partially supported by grants from National Science and Technology Major Project under Grant (No. 2023ZD0121103), the National Natural Science Foundation of China (No. 62376086, U23B2031, U23B2031, 721881011).

\bibliography{7-reference}

\clearpage

\appendix
\section{Datasets and Additional Experiments on NLG}
\label{s:nlg}
\noindent \textbf{Datasets. } 
We utilized GLUE benchmark~\citep{wang2018glue} to evaluate the model performance. 
GLUE covers multiple tasks of paraphrase detection (MRPC, QQP), sentiment classification (SST-2), natural language inference (MNLI, RTE, QNLI), and linguistic acceptability (CoLA). 
Following previous work~\citep{zhangrevisiting}, for those datasets with fewer than $10,000$ samples (i.e., RTE, MRPC, STS-B, CoLA), we split the original validation set into new validation and test sets equally. 
For others, we randomly select $1,000$ examples from training set as the validation set, and use original validation sets as test sets. Additionally, we included the BoolQ~\citep{clark2019boolq}, PIQA~\citep{bisk2020piqa}, OBQA~\citep{mihaylov2018can}, and ARC~\citep{clark2018think} datasets to assess the model's performance in commonsense reasoning tasks. These datasets provide a variety of challenges that require understanding of everyday scenarios and logical reasoning.
Moreover, we select SciTail~\citep{khot2018scitail}, BoolQ~\citep{clark2019boolq}, and CB~\citep{de2019commitmentbank} datasets to evaluate model robustness and generalization capabilities in few-shot learning scenarios.

To further validate the effectiveness of our method, we conducted experiments on natural language generation (NLG) tasks using three datasets: DART, E2E, and WebNLG. DART focuses on generating text from structured data, E2E involves generating restaurant descriptions from key attributes, and WebNLG is designed for generating text from knowledge graph triples. As shown in Table \ref{tab:nlg_experiments}, none of the methods outperform fine-tuning (FT) on NLG tasks, and LoRA shows a significant performance drop. This indicates that using a fixed rank for training all tasks is suboptimal. In contrast, our method achieves performance comparable to FT. This is attributed to our method's ability to allocate an appropriate rank for different tasks efficiently.

\noindent \textbf{Evaluation Setup. } 
For GLUE benchmark and commonsense reasoning tasks, we selected the checkpoint with the highest average performance on validation set. 
For few-shot learning, we performed training and testing under each shot setting using 5 random seeds. 
Then, we reported the average performance for a fair and robust estimation and comparison. 

\begin{table}[t!]
	\centering
	\begin{tabular}{cccccc}
		\toprule
		\textbf{Method} & \textbf{DART} & \textbf{E2E} & \textbf{WebNLG} & \textbf{AVG} \\
		\midrule
		FT & \textbf{46.1} & \underline{61.4} & 44.2 & \textbf{50.6} \\
		\textbf{LoRA}$_{r=8}$ & 43.2 & 60.6 & 43.8 & 49.2 \\
		\textbf{LoRA}$_{r=16}$ & 44.6 & 60.8 & 44.3 & 49.9 \\
		MultiLoRA & 44.0 & 61.3 & 44.9 &  50.1 \\
		MixLoRA & 44.3 & 60.9 & \textbf{45.3} & 50.2 \\
		MoRE & \underline{45.0} & \textbf{61.5} & \underline{45.1} & \underline{50.5} \\
		\bottomrule
	\end{tabular}
	\caption{Model performance on NLG tasks}
	\label{tab:nlg_experiments}
\end{table}


\section{Detailed Calculation of Parameter Counts}
\textbf{Parameter Efficiency. }
\label{s:parameter-count}
To analyze the model complexity, we give the number of tuning parameters of different \la-based methods and report results in Table~\ref{tab:methods_params}. 
The notation explanations are as follows: 
$\{L, r, (m, d), n, T, h\}$ refer to model layers, LoRA rank, model dimensions, parallel LoRA module number, task number, and task embedding dimension.
Compared with tuning parameter size of \la,  the added parameter number of \name~is $6Lh(r+T)$, including the extra task embeddings ($Th$ for a single \la~module) and adaptive rank selector ($rh$ for a single \la~module). 
Compared with MultiLoRA and MixLoRA which use parallel module design to tackle multi-task learning, \name~is more efficient. 
Moreover, once our task embedding and gate modules are trained, we can construct a mapping from tasks to experts, which allows us to avoid the repeated computation of the task embedding and gate modules during inference, thereby reducing the parameter count to be consistent with $LoRA$. 
This is also the reason why \name~achieves impressive performance in multi-task scenarios without too many fine-tuning parameters. 

\paragraph{LoRA parameters:} LoRA employs matraix A and B to introduce low-rank adaptations in both the attention layers (q, k, v, o) and the feed-forward network (FFN) layers ($w_i$, $w_o$) of the T5-base model. Each LoRA layer has $r(m+d)$ paramters. The total number of parameters for LoRA with L transformer layers is $6Lr(m+d)$.

\paragraph{MultiLoRA parameters:} MultiLoRA employs parallel LoRA models for training, so its parameter count is $n$ times that of vanilla LoRA, where $n$ is the number of parallel LoRA modules. Additionally, MultiLoRA modifies the scaling factors to be learnable parameters (with parameter count $d$). Therefore, the total number of parameters is $6nLr(m+d)+6Ld$.

\paragraph{MixLoRA parameters:} MixLoRA only employs parallel expert LoRA modules in the FFN layers and uses a gating module (with parameter count $nm$) to select the appropriate LoRA expert. Therefore, the total number of parameters is $2nLr(m+d)+2Lnm$.

\paragraph{MOELoRA Parameters:} MOELoRA utilizes parallel LoRA models with a rank of $r/n$ and incorporates a task embedding module to represent each task (with a parameter count of $Th$). Additionally, it employs a gating module (with a parameter count of $nh$) to compute the weights for each LoRA. Therefore, the total number of parameters is given by $6Lr(m+d) + 6Lh(n+T)$.

\paragraph{MoRE parameters:} Our proposed MoRE 
employs the same LoRA modules as vanilla LoRA, but treats LoRA modules with different ranks as experts, thereby introducing an additional gating module (with parameter count $rh$). To better adapt to different tasks, we also introduce a task embedding module (with parameter count $Th$, where h is the hidden dimension). Therefore, the total number of parameters is $6Lr(m+d)+6Lh(r+T)$. In the GLUE dataset, \(T=8\) is consistent with $r=8$. If the hidden dimension is set to be the same as $d$, then the parameter count is $12Lr(m+d)$, which is exactly the same as the parameter count  with $LoRA_{r=16}$. Compared to MultiLoRA and MixLoRA, we do not use a parallel module design, so there is no parameter $n$ that leads to a parameter count far exceeding that of LoRA. Furthermore, once our task embedding and gate modules are trained, we can construct a mapping from tasks to experts. This allows us to avoid the repeated computation of the task embedding and gate modules during inference, thereby reducing the parameter count to be consistent with $LoRA_{r=8}$.

\section{Additional Visualization of Task Embeddings}
\label{s:add_vis}
\begin{figure*}[htbp]
	\centering
	\includegraphics[width=0.90\textwidth]{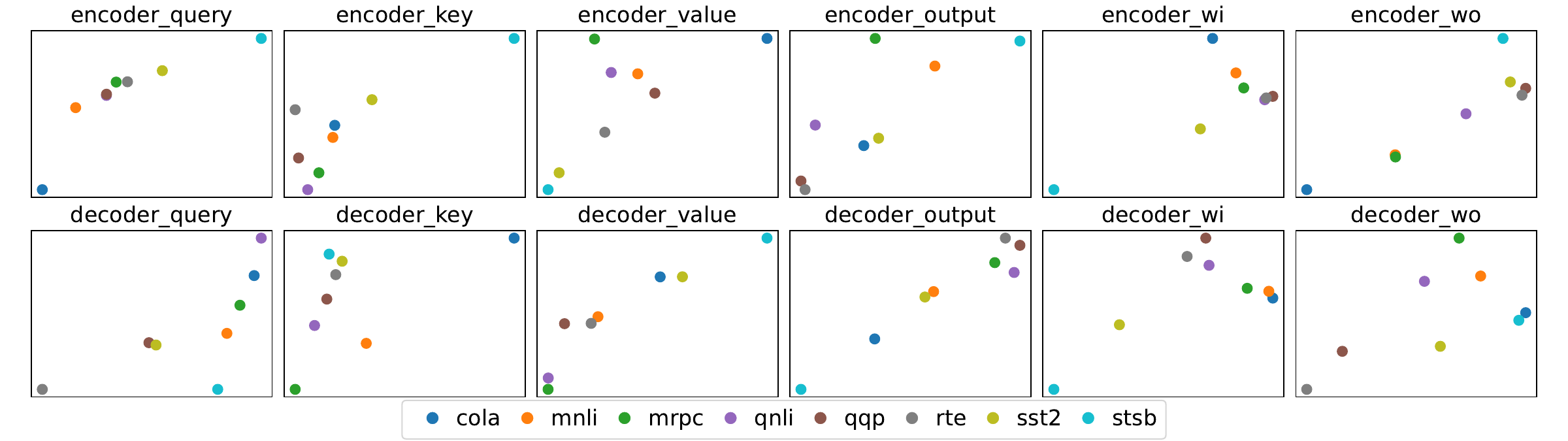}
	\caption{Visualization of Task Embeddings in Layer 1.}
	\label{fig:layer1}
\end{figure*}

\begin{figure*}[htbp]
	\centering
	\includegraphics[width=0.90\textwidth]{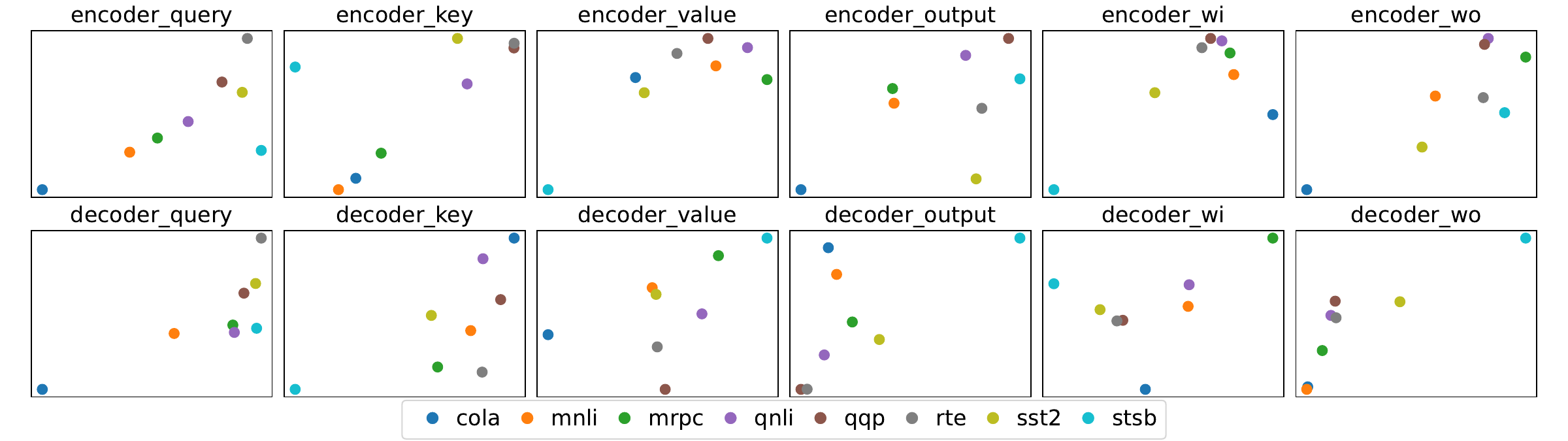}
	\caption{Visualization of Task Embeddings in Layer 6.}
	\label{fig:layer6}
\end{figure*}

\begin{figure*}[htbp]
	\centering
	\includegraphics[width=0.90\textwidth]{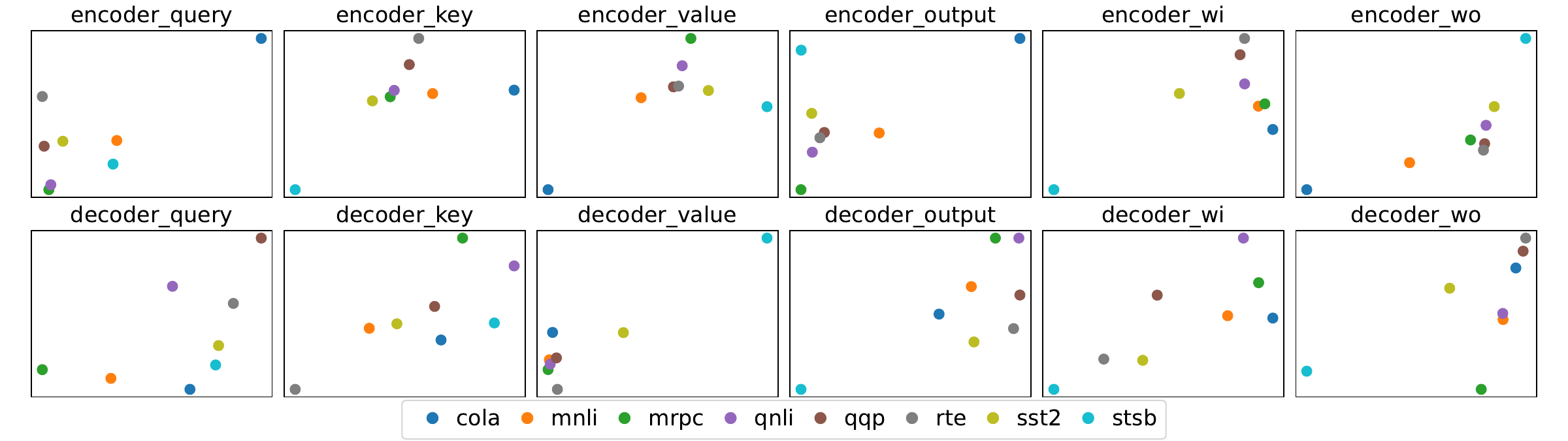}
	\caption{Visualization of Task Embeddings in Layer 12.}
	\label{fig:layer12}
\end{figure*}

Further analysis of task embeddings is presented in Figures \ref{fig:layer1}-\ref{fig:layer12}. These figures reveal that the patterns observed in other layers and modules of the model are consistent with those reported in the main text. Notably, stronger clustering is observed in the $w_i$ and $w_o$ layers. This enhanced clustering may be attributed to the feed-forward network (FFN) layers' ability to capture shared information underlying different tasks more effectively.

\end{document}